%% file: main.tex
\documentclass{article}
\usepackage{ijcai20}

\usepackage{times}
\usepackage{soul}
\usepackage{url}
\usepackage[hidelinks]{hyperref}
\usepackage[utf8]{inputenc}
\usepackage[small]{caption}
\usepackage{graphicx}
\usepackage{amsmath}
\usepackage{amsthm}
\usepackage{booktabs}
\usepackage{algorithm}
\usepackage{algorithmic}
\algsetup{linenosize=\small}
\usepackage[dvipsnames]{xcolor}
\usepackage{mysymbols}
\usepackage{cleveref}
\usepackage{wrapfig}
\usepackage{subfig}
\usepackage{comment}
\usepackage{etoolbox}
\newtoggle{arxiv}
\toggletrue{arxiv} %
\input{math_commands.tex}

\input{macros.tex}

\title{IR-VIC: Unsupervised Discovery of Sub-goals for Transfer in RL}

\author{
Nirbhay Modhe$^1$
\and
Prithvijit Chattopadhyay$^1$\and
Mohit Sharma$^{1}$\and
Abhishek Das$^{1}$\and\\ %
Devi Parikh$^{1,2}$\and
Dhruv Batra$^{1,2}$\And
Ramakrishna Vedantam$^2$
\affiliations
$^1$Georgia Institute of Technology\\
$^2$Facebook AI Research
\emails
\{nirbhaym,prithvijit3,mohit.sharma,abhshkdz,parikh,dbatra\}@gatech.edu,
ramav@fb.com
}

\begin{document}

\maketitle

\begin{abstract}
\rama{We propose a novel framework to identify sub-goals useful
for exploration in sequential
decision making tasks under partial observability.
We utilize the variational intrinsic control framework (Gregor et.al., 2016) which maximizes
empowerment -- the ability to reliably reach a diverse set of states
and show how to identify sub-goals as states with high \emph{necessary} option information
through an information theoretic regularizer. Despite being discovered
without explicit goal supervision, our sub-goals provide better
exploration and sample complexity 
on challenging grid-world navigation tasks
compared to 
\nirbhay{supervised counterparts in prior work.}
}
\end{abstract}

\section{Introduction}\label{sec:intro}
\input{new_sections/intro.tex}

\section{Methods}\label{sec:methods}

\input{new_sections/methods.tex}

\section{Experiments}\label{sec:results}
\input{new_sections/results.tex}

\section{Related Work}\label{sec:related}
\input{new_sections/related_work.tex}
\section{Conclusion}\label{sec:discussion}
\input{new_sections/conclusion.tex}

\bibliographystyle{named}
\bibliography{new_bib}

\iftoggle{arxiv}{
\appendix
\section{Appendix}

\input{new_sections/appendix.tex}

}{}
\end{document}

%% file: math_commands.tex
\usepackage{amsmath,amsfonts,bm}

\def\Figref#1{Figure~\ref{#1}}

\def\eqref#1{equation~\ref{#1}}
\def\Eqref#1{Equation~\ref{#1}}

\def\1{\bm{1}}

\def\vs{{\bm{s}}}

\DeclareMathAlphabet{\mathsfit}{\encodingdefault}{\sfdefault}{m}{sl}
\SetMathAlphabet{\mathsfit}{bold}{\encodingdefault}{\sfdefault}{bx}{n}

\newcommand{\E}{\mathbb{E}}

\newcommand{\KL}{D_{\mathrm{KL}}}

\makeatletter
\newcommand{\subalign}[1]{%
  \vcenter{%
    \Let@ \restore@math@cr \default@tag
    \baselineskip\fontdimen10 \scriptfont\tw@
    \advance\baselineskip\fontdimen12 \scriptfont\tw@
    \lineskip\thr@@\fontdimen8 \scriptfont\thr@@
    \lineskiplimit\lineskip
    \ialign{\hfil$\m@th\scriptstyle##$&$\m@th\scriptstyle{}##$\hfil\crcr
      #1\crcr
    }%
  }%
}
\makeatother

%% file: macros.tex
\newcommand{\mohit}{\textcolor{black}}

\newcommand{\nmcr}{\textcolor{black}}
\newcommand{\nirbhay}{\textcolor{black}}
\newcommand{\nmrebuttal}{\textcolor{black}}

\newcommand{\prithvi}{\textcolor{black}}

\newcommand{\rama}{\textcolor{black}}

\newcommand{\fourroom}{\texttt{4-Room}\xspace}
\newcommand{\maze}{\texttt{maze}\xspace}
\newcommand{\ibot}{InfoBot\xspace}
\newcommand{\irvic}{IR-VIC\xspace}
\newcommand{\diayn}{DIAYN\xspace}
\newcommand{\vic}{VIC\xspace}
\newcommand{\decs}{decision states\xspace}

\newcommand{\uatext}[1]{\noalign{\noindent\textrm{{#1}}}}

\newtheorem{theorem}{Theorem}[section]

\newtheorem{lemma}[theorem]{Lemma}

\makeatletter
\DeclareRobustCommand\onedot{\futurelet\@let@token\@onedot}
\def\@onedot{\ifx\@let@token.\else.\null\fi\xspace}

\def\eg{\emph{e.g}\onedot} 
\def\ie{\emph{i.e}\onedot} 
\def\st{s.t\onedot} 
 
 \def\vs{\emph{vs}\onedot}
 
\makeatother

\newcommand{\myvecsym}[1]{\boldsymbol{#1}}

\newcommand{\vtau}{\myvecsym{\tau}}

\newcommand{\eat}[1]{}
\newcommand{\remark}[2]{} %
\newcommand{\replace}[2]{} %

\iftoggle{arxiv}{
\setlength\marginparsep{5pt}
\setlength\marginparwidth{\dimexpr1in+\hoffset+\oddsidemargin-\marginparsep*2}
}{}

\newenvironment{packed_itemize}{
	\begin{list}{\labelitemi}{\leftmargin=2em}
	    \iftoggle{arxiv}{
		\setlength{\itemsep}{0pt}
		\setlength{\parskip}{0pt}
		\setlength{\parsep}{0pt}
		}{}
	}{\end{list}}

%% file: new_sections/intro.tex
A common approach in reinforcement learning (RL) is
to decompose \rama{an original decision making problem} into a set of
simpler \rama{decision making problems} -- each \rama{terminating into an
identified sub-goal}.
\rama{Beyond such a decomposition or abstraction being evident in humans
(\eg adding salt is a sub-goal in the process of cooking a dish)~\citep{hayes1979cognitive},
sub-goal identification is also useful from a practical perspective of constructing
policies that transfer to novel tasks (\eg adding salt is a useful sub-goal across
a large number of dishes one might want to cook, corresponding to different `end' goals).}

However,
identifying sub-goals that can 
accelerate learning
while also being re-usable across tasks or environments 
is a challenge in itself. Constructing
such sub-goals often requires knowledge of the task
structure (supervision) 
and may fail in cases where 1) dense rewards are absent \citep{pathak2017curiosity}, 
2) rewards require extensive hand engineering and domain knowledge (hard to scale),
and 3) where the notion of reward may not be obvious \citep{lillicrap2015continuous}. 
In this work, we demonstrate a method for identifying 
sub-goals in an ``unsupervised'' manner -- \rama{without any external
rewards or \rama{goals}}.
We show that our sub-goals generalise to novel \rama{partially
observed} environments and \rama{goal-driven} tasks,
leading to comparable (or better) performance \rama{(via.
better exploration)} on
downstream tasks compared to prior work on goal-driven 
sub-goals \citep{goyal2019infobot}.

We study sub-goals in the framework of
quantifying the minimum information necessary 
for taking actions by an agent. \citet{van2011grounding} have shown that in the 
presence of an external goal, the minimum goal information required by an agent 
for taking an action is a \nirbhay{useful measure of sub-goal states.} 
\citet{goyal2019infobot} demonstrate that \nirbhay{for action $A$, state $S$ and a goal $G$,}
such sub-goals can be efficiently learnt by imposing 
a bottleneck on the information $I(A, G | S)$.
We show that replacing the goal with an intrinsic objective admits a 
strategy for discovery of sub-goals in a completely 
unsupervised
manner.

\begin{figure}[t!]
    \centering
    \includegraphics[width=0.45\textwidth]{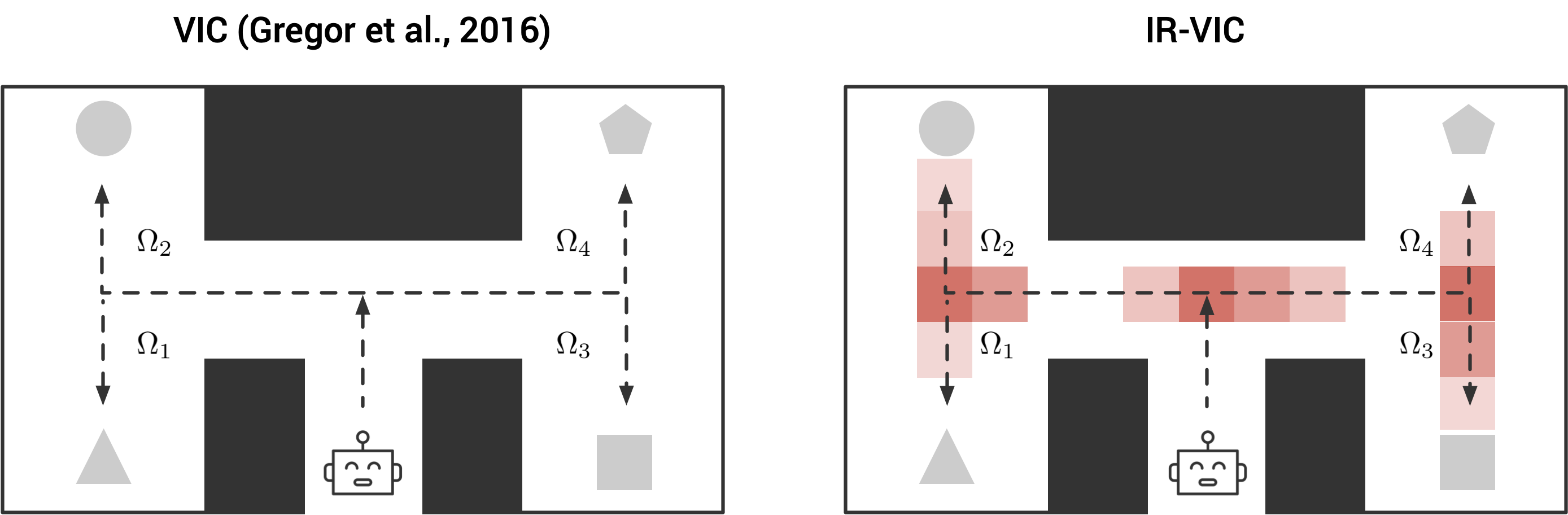}
    \caption{
    Left: The VIC framework 
    ~\protect\cite{gregor2016variational}
    in a navigation context: an agent learns high-level macro-actions (or options)
    to reach different states in an environment reliably without
    any extrinsic reward.
    Right: \nirbhay{IR}-VIC identifies 
    \nirbhay{
    sub-goals as states where necessary option information is high 
    (darker shades of red) for an empowered agent.
    }
    Identification of \nirbhay{unsupervised sub-goals} leads to improved
    transfer to novel environments.
    }
    \label{fig:teaser_fig}
\end{figure}

\nirbhay{Our choice of intrinsic objective is}
the Variational Intrinsic Control (\vic) formulation \citep{gregor2016variational} 
to learn options $\Omega$ that maximize 
the mutual information $I(S_f, \Omega)$, \nirbhay{referred to as empowerment}, 
where  $S_f$ is the final state in a trajectory ~\citep{Salge2013-hr}.
To see why this maximizes empowerment,
notice that $I(S_f, \Omega) = H(S_f) - H(S_f | \Omega)$,
where $H(.)$ denotes entropy. Thus, empowerment maximizes the diversity in
final states $S_f$ while learning options highly predictive
of $S_f$.
\nirbhay{We demonstrate that}
by limiting the information the agent uses about the selected option
$\Omega$ \nirbhay{while maximizing empowerment,} 
a sparse set of states emerge where the necessary option
information $I(\Omega, A|S)$ is high -- we interpret these states
as our unsupervised sub-goals. 
We call our approach Information Regularized VIC (IR-VIC).
Although IR-VIC is similar in spirit to ~\cite{goyal2019infobot,polani2006relevant},
it is important to note that we use latent options $\Omega$ instead of external goals --
removing any dependence on the task-structure.
To summarize our contributions,
\begin{packed_itemize}
\item We propose Information Regularized VIC (\irvic), a novel framework to identify sub-goals
    in a task-agnostic manner, \rama{by regularizing relevant option information}.
\item \rama{Theoretically, we show that the proposed objective is a sandwich 
    bound on the empowerment $I(\Omega, S_f)$ -- this is the only useful upper bound we 
    are aware of}.
\item We show that our \nirbhay{sub-goals are}
    transferable and lead to improved sample-efficiency on goal-driven tasks
    in novel, \rama{partially-observable} environments.
    On a challenging grid-world navigation task,
    our method outperforms (a re-implementation of)~\cite{goyal2019infobot}.
\end{packed_itemize}

%% file: new_sections/methods.tex
\subsection{Notation}
\label{sec:notation}
We consider a Partially Observable 
Markov Decision Process (POMDP),
defined by the tuple $(\mathcal{S}, \nirbhay{\mathcal{X},} \mathcal{A}, \mathcal{P}, r)$,
$s \in \mathcal{S}$ is the state,
$x \in \mathcal{X}$ is the partial observation of the state
and $a\in\mathcal{A}$ is an action from a discrete action space.
$\mathcal{P} : \mathcal{S} \times \mathcal{S} \times \mathcal{A}$ 
denotes an unknown
transition function,
representing $p\left(s_{t+1} | s_{t}, a_{t}\right): s_t, s_{t+1} \in \mathcal{S}, A_t \in \mathcal{A}$.
Both~\vic and~\irvic
\nirbhay{initially} train an option ($\Omega$) conditioned policy $\pi(a_t| \omega, \nirbhay{x_t})$, where
$\omega \in \{1,\cdots, |\Omega|\}$. \nirbhay{\rama{During transfer, all approaches
(including baselines)
train
a goal-conditioned policy }$\pi(a_t| \nirbhay{x_t}, g_t)$ where $g_t$ is the
goal information at time $t$.}
Following standard practice~\citep{Cover91},
we denote random variables in uppercase ($\Omega$), and
items from the sample space of random variables in lowercase ($\omega$). 

\subsection{Variational Intrinsic Control (\vic)}
\label{sec:vic_desc}
\vic maximizes the mutual information between
options $\Omega$ and the final (option termination)
state $S_f$ given $s_0$, 
\ie $I(S_f, \Omega \mid S_0=s_0)$, which encourages the agent to 
learn options that reliably reach a diverse set of states. 
This objective is estimated
by sampling an option from a prior at the
start of a trajectory, following it until termination,
and inferring the sampled option given the final and initial states.
Informally, VIC maximizes the empowerment for
an agent, \ie its internal options $\Omega$ 
have a high degree of correspondence to the
states of the world $S_f$ that it can reach. \vic
formulates a variational
lower bound on this mutual information. Specifically,
let $p(\omega \mid s_0)=p(\omega)$ be a prior on options
\nmrebuttal{(we keep the prior fixed as per \cite{eysenbach2018diversity}),}
$p^J(s_f\mid \omega, s_0)$ 
\nmrebuttal{is defined as}
the
(unknown) terminal state distribution
achieved when
executing the policy $\pi(a_t\mid \omega, s_t)$,
and $q_{\nu}(\omega \mid s_f, s_0)$ denote
a (parameterized) variational approximation to
the true posterior on options given
$S_f$ and $S_0$. Then:
\begin{align}\label{eqn:vic_objective}
    &I(\Omega, S_f\mid S_0=s_0) \notag\\
    &\ge \E_{
        \subalign{\Omega &\sim p(\omega)\\
        S_f &\sim p^J(s_f \mid \Omega, S_0=s_0)}
    }
    \Bigg[ 
        \log \frac{q_{\nu}(\Omega \mid S_f, S_0 = s_0)}{p(\Omega)}
    \Bigg]\\
    &= \mathcal{J}_{VIC}(\Omega, S_f; s_0)\notag
\end{align}

\begin{figure}[t]
	\centering
	\includegraphics[width=0.48\textwidth]{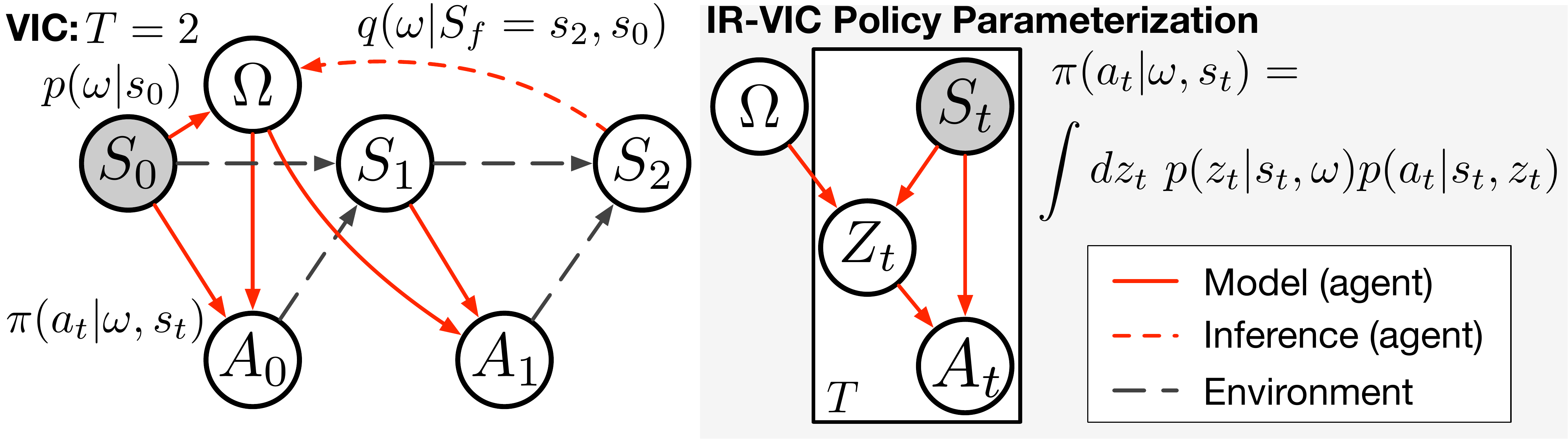}
    \caption{
	{\textbf{Illustration of~\vic for 2 timesteps.} 
	\textbf{L:} Given a start state $S_0$, VIC samples option
	$\omega$ and follows policy $\pi(a_t \mid \Omega=\omega, s_t)$
	and infers
	$\Omega$ from the terminating
	state ($S_2$), optimizing a lower bound on $I(S_2, \Omega\mid S_0)$.
	\textbf{R:} \irvic considers a particular parameterization
	of $\pi$ and imposes a bottleneck on $I(A_t, \Omega| S_t)$.}}
	\label{fig:graphical_model}
\end{figure}

\subsection{Information Regularized VIC (IR-VIC)}
\label{sec:ds_vic}
\nirbhay{We identify sub-goals as 
states where the necessary option
information required for deciding actions is high.
} 
Formally, this
means that at every timestep $t$ in the trajectory, we
minimize the mutual information
$I(\Omega, A_t| S_t, S_0=s)$ 
\nirbhay{, resulting in a sparse set of states 
where this mutual information
remains high 
despite the minimization}.
Intuitively, this means that
on average (across different options),
\nirbhay{these states have higher relevant option
information that other states}
(\eg the regions with darker shades of red in \Figref{fig:teaser_fig}).
Overall, our objective is to maximize:
\begin{equation}\label{eqn:usd_objective}
\mathcal{J}_{VIC}(\Omega, S_f; s_0)
- \beta \sum_t \; I(\Omega, A_t \mid S_t, S_0=s_0)
\end{equation}
where $\beta>0$ is a trade-off parameter.
Thus, this is saying
that one wants options $\Omega$ which allow the agent to have
a high empowerment, while utilizing the least relevant
\mohit{option} information at each step. 

Interestingly,~\Eqref{eqn:usd_objective} has a clear,
principled interpretation
in terms of the empowerment $I(\Omega, S_f| S_0)$ from the~\vic
model. We state the following lemma 
\iftoggle{arxiv}{
(proof in \Cref{appendix:lemma_proof}):
}{
(\nirbhay{which follows from
recursively applying the chain rule of mutual information
and the data-processing inequality \citep{Cover91}}):
}

\begin{lemma}\label{lemma:sandwich_bound}
Let $A_t$ be the action random variable at timestep
$t$ and state $S_t$
following an option-conditioned policy $\pi(a_t | s_t, \omega)$.
Then, $I(\Omega, A_t | S_t, S_0)$ \ie the conditional mutual
information between the option $\Omega$ and action
$A_t$ when summed over all timesteps in the trajectory, upper bounds the
conditional mutual information $I(\Omega, S_f | S_0)$ between $\Omega$ and
the final state $S_f$ -- namely the empowerment as
defined by~\citet{gregor2016variational}:
\begin{equation}\label{lemma_eqn}
    I(\Omega, S_f | S_0) \leq \sum_{t=1}^{f} I(\Omega, A_{t} | S_t, S_0) = \mathcal{U}_{DS}(\vtau, \Omega, S_0)
\end{equation}
\end{lemma}

\paragraph{\textbf{Implications.}} 
With this lens, one can view
the optimization
problem in~\Eqref{eqn:usd_objective} as a Lagrangian relaxation of
the following constrained optimization problem:
\begin{equation}
    \max\; \mathcal{J}_{VIC} \quad \st{}\ \text{ } \mathcal{U}_{DS} \leq R
\end{equation}
where $R > 0$ is a constant.
\rama{While upper bounding the empowerment does not directly
imply one will find 
\nirbhay{useful sub-goals}
(meaning it is the structure
of the decomposition~\cref{lemma_eqn} that is more relevant than the fact that
it is an upper bound), this bound might be of interest more generally
for representation learning~\citep{achiam2018variational,gregor2016variational}.}
\rama{Targeting specific values for the upper bound $R$ can potentially allow
us to control how `abstract' or invariant the latent option representation
is relative to the states $S_f$, leading to solutions that say,
neglect unnecessary information in the state representation to allow
better generalization. 
Note that most approaches currently limit the abstraction
by constraining the number of discrete options, which (usually) imposes
an upper bound on 
$I(\Omega, S_f) = H(\Omega) - H(\Omega| S_f)$, since $H(\Omega) \ge H(\Omega| S_f)$ and
$H\ge 0$ in the discrete case.
However, this does not hold for the continuous case, where this result
might be more useful. Investigating this is beyond the scope of this
current paper, however, as our central aim is to identify
useful sub-goals, and not to scale the \vic framework to continuous options.}

\subsection{Algorithmic Details}

\subsubsection{Upper Bounds for $I(\Omega,A_t\mid S_t,S_0)$}
Inspired by~\ibot~\citep{goyal2019infobot}, we \nirbhay{bottleneck} the information in a
statistic $Z_t$ of the state $S_t$ and option $\Omega$ used to parameterize
the policy $\pi(A_t\mid \Omega, S_t)$ (\cref{fig:graphical_model} right). This is
justified by the
the data-processing inequality~\citep{Cover91} for the markov chain
$\Omega, S_t \leftrightarrow Z_t \leftrightarrow A_t$, which implies
$I(\Omega, A_t\mid S_t, S_0) \leq I(\Omega, Z_t\mid S_t, S_0)$.
We can then obtain the following upper bound on $I(\Omega, Z_t\mid S_t, S_0)$:~
\footnote{Similar to VIC, $p^J$ here denotes the (unknown) state distribution at time $t$ from which
we can draw samples when we execute a policy. We then assume a variational approximation $q(z_t)$ (fixed to be a unit gaussian)
for $p(z_t | S_t)$. Using the fact that $D_{\mathtt{KL}}(p(z_t | s_t)||q(z_t)) \geq 0$ we get the derived upper bound.}
\begin{align}
&I(\Omega, Z_t\mid S_t, S_0=s) \notag \\
&\le \E_{
    \subalign{
        \Omega &\sim p(\omega) \\
        S_t &\sim p^J(s_t\mid \Omega, S_0=s) \\
        Z_t &\sim p(z_t\mid S_t, \Omega)}}
\left[
    \log \frac{p(Z_t\mid \Omega, S_t)}{q(Z_t)}
\right]
\label{eq:usd_local_bound}
\end{align}
where $q(z_t)$ is a
fixed variational approximation (set to $\mathcal{N}(0, \mathtt{I})$
as in \ibot), and $p_{\phi}(z_t\mid\omega, s_t)$
is a parameterized encoder. As explained in~\cref{sec:intro},
the key difference between~\cref{eq:usd_local_bound}
and~\ibot is that they construct upper bounds on
$I(G, A_t \mid S_t, S_0)$ \nirbhay{using information about the goal $G$}, 
while we bottleneck the option-information. \rama{One could use the
DIAYN objective~\citep{eysenbach2018diversity}
(see more below under related objectives) which also
has a $I(A_t, \Omega| S_t)$ term, and directly bottleneck
the action-option mutual information instead of~\cref{eq:usd_local_bound}, but
we found that directly imposing this bottleneck often hurt convergence in practice.}

We can compute a Monte Carlo estimate of~\Eqref{eq:usd_local_bound}
by first sampling an option $\omega$ at $s_0$ and then
keeping track of all states visited in trajectory $\vtau$.
In addition to the~\vic term and our bottleneck regularizer,
we also include the entropy of the policy over the actions
(maximum-entropy RL~\citep{ziebart2008maximum}) as a bonus to encourage sufficient
exploration. \nmcr{We fix the coefficient for maximum-entropy, $\alpha = 10^{-3}$ 
which consistently works
well for our approach as well as baselines.} Overall, the \irvic objective is:
\begin{align}
&\max_{\theta, \phi, \nu} \tilde{J}(\theta, \phi, \nu)
= \E_{
    \subalign{
        \Omega &\sim p(\omega)\\
        \vtau &\sim \rama{\pi(\cdot\mid\omega, S_0)}\\
	    Z_t &\sim p_{\phi}(z_t\mid S_t, \Omega)\\
	}
}\notag
\Bigg[ 
    \log \frac{q_{\nu}(\Omega \mid S_f, S_0)}{p(\Omega)} \\
    &- \sum_{t=0}^{f-1}\Big(\beta \log \frac{p_{\phi}(Z_t \mid S_t, \Omega)}{q(Z_t)} +
    \alpha  \log \pi_{\theta}(A_t \mid S_t, Z_t) \Big)
\Bigg]
\label{eq:usd_final_obj}
\end{align}
where $\theta, \phi$ and $\nu$ are the parameters of the policy,
latent variable decoder and the option inference network respectively.
The first term in the objective \mohit{promotes high empowerment} 
while learning options; the second term ensures
\textit{minimality} in using the options sampled to \nirbhay{take actions} and the third provides an
incentive for \textit{exploration}.

\subsubsection{Related Objectives}
\diayn \citet{eysenbach2018diversity} attempts to learn
skills (similar to options) which can control the states visited by agents
while ensuring that 
\nirbhay{all visited states, as opposed to termination states,}
are used to distinguish skills.
Thus, for an option $\Omega$ and every state $S_t$ in a trajectory,
they maximize $\sum_t I(\Omega, S_t) - I(A_t, \Omega| S_t) + H(A_t| S_t)$,
as opposed to $I(\Omega, S_f) - \beta \sum_t I(A_t, \Omega| S_t) + H(A_t | S_t)$
in our objective. With the sum over all timesteps for $I(\Omega, S_t)$,
the bound in~\cref{lemma:sandwich_bound} no
longer holds true, which also means that there is
no principled reason (unlike our model) to scale the second term with
$\beta$.
The most closely related work to ours is~\ibot~\citep{goyal2019infobot}, which maximizes
$\sum_t R(t) - \beta \rama{I(Z_t, G| S_t)}$
for a goal ($G$) conditioned policy
$\pi(a_t| S_t, G)$. \nirbhay{They define states where \rama{$I(Z_t, G| S_t)$}
is high despite the bottleneck as ``\decs''.}
The key difference
is that~\ibot requires extrinsic rewards \nirbhay{in order to identify
goal-conditioned decision states},
while our work is strictly more general and scales
in the absence of extrinsic rewards.

Further, in context of both these works, our work provides a
principled connection between action-option information reqularization
$I(A_t, \Omega| S_t)$ and empowerment of an agent. The tools from~\Cref{lemma:sandwich_bound}
might be useful for analysing these previous objectives which
both employ this technique.

\subsection{Transfer to Goal-Driven Tasks}

\nirbhay{In order to transfer sub-goals to novel 
environments,} \citet{goyal2019infobot} pretrain
their model to identify their \nirbhay{goal-conditioned} \decs, 
and then study if identifying similar states in new environments can 
improve exploration when training a new policy
$\pi_{\gamma}(a|s, g)$ from scratch. Given an environment
with reward $R_e(t)$, goal $G$, $\kappa > 0$,
and state visitation count $c(S_t)$,
their reward is:
\begin{equation}\label{eq:bonus_reward_infobot}
R_t = R_e(t) + \frac{\kappa}{\sqrt{c(S_t)}} \underbrace{I(G, Z_t | S_t)}_{\text{Pretrained, Frozen}}
\end{equation}

The count-based reward
decays with square root of
 $c(S_t)$ to encourage the model to explore novel states, and the
mutual information between goal $G$ and bottleneck variable $Z_t$
\nirbhay{is a smooth measure of whether a state is a sub-goal,}
and is multiplied with the
exploration bonus to encourage visitation of state where 
\nirbhay{this measure is high}.

\begin{figure}[t]
	\centering
	\includegraphics[width=0.45\textwidth]{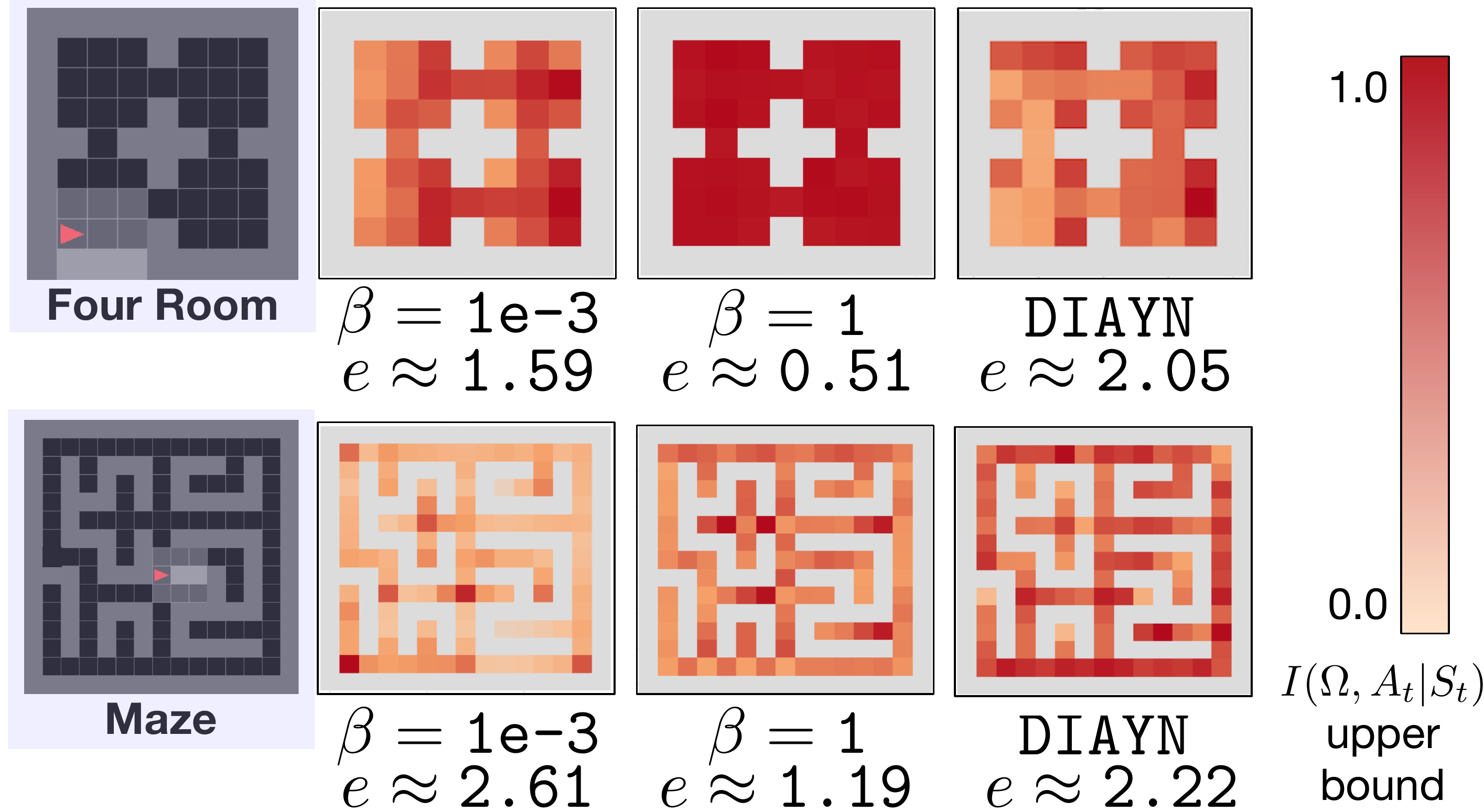}
	\caption{
	\nirbhay{Heatmaps of necessary option information 
	$I(\Omega, Z_t| S_t, S_0)$ (normalized to [0,1]) 
	at visited states}
	on environments -- \texttt{4-Room} (top) and \texttt{maze} (bottom).
	First column depicts environment layout, second and third show results 
	for \irvic, $\beta=1e^{-3}$ and $\beta=1$ respectively, 
	and the fourth column shows \diayn.
	$e$-values show computed lower bounds (with Eq.~\ref{eqn:vic_objective}) on empowerment
	in nats.
	}
	\label{fig:prelim_env_dec_states}
\end{figure}

We use an almost identical setup, replacing their decision-state term
from supervised pretraining with \nirbhay{necessary option information from \irvic} pretraining:

\begin{equation}\label{eq:bonus_reward}
R_t = R_e(t) + \frac{\kappa}{\sqrt{c(S_t)}} \underbrace{I(\Omega, Z_t | S_t, S_0)}_{\text{Pretrained, Frozen}}
\end{equation}

\noindent
$I(\cdot)$ is computed with~\cref{eq:usd_local_bound} with
a frozen parameterized encoder $p(z_t\mid \omega, s_t)$
during transfer. Thus, we incentivize
visitation of states where necessary option information
is high.

\iftoggle{arxiv}{}{
\par \noindent
\paragraph{\textbf{Algorithm.}} In summary, \irvic consists of
two major phases -- (1) Unsupervised Discovery and (2) Transfer. In phase (1), we optimize Eqn.~\ref{eq:usd_final_obj} based on unrolled
trajectories conditioned on the initially sampled option to update the parameterized encoder $p_{\phi}(z_t|\omega, x_t)$, option-conditioned policy $\pi_{\theta}(a_t| \omega, x_t)$ and the option inference network $q_{\nu}(\omega | s_0, s_f)$. In phase (2), we freeze the pretrained encoder
$p_{\phi}(z_t|\omega, x_t)$ and use it as an added
exploration incentive to learn the goal conditioned
policy $\pi_{\gamma}(a_t|x_t, g)$ (see Eqn.~\ref{eq:bonus_reward}).
}

\subsection{IR-VIC for Transfer}
\paragraph{\textbf{Options with partial observability.}}
\nirbhay{The methods we have described so far have assumed
the true state $s\in\mathcal{S}$ to be known --} 
\rama{the VIC framework with explicit options
has only been shown to work in fully 
obervable MDPs~\citep{gregor2016variational,eysenbach2018diversity}. 
However, since we are primarily interested improved exploration in downstream
partially-observable tasks, we adapt the VIC framework to only use
partially-observable information for the parts that we use during transfer.
We design our policy (including the encoder $p(Z_t \mid \Omega, S_t)$ 
used for computing the reward bonus $I(\Omega, Z_t\mid S_t, S_0)$ during transfer)
to take as input partial observations $x\in\mathcal{X}$
while allowing the option inference networks (of \irvic and \diayn)
to take as input the global (\texttt{x, y}) coordinates of the agent 
(assuming access to the true state $s\in\mathcal{S}$.
\nirbhay{Note that this privileged information 
is made available for a single environment in order to 
discover sub-goals transerable to multiple novel environments 
(whereas supervised methods such as InfoBot \citep{goyal2019infobot} 
require global (\texttt{x, y}) coordinates as goal information
across all training environments).
}}
\par
\noindent
\nirbhay{
\paragraph{\textbf{Preventing option information leak.}}
\nmcr{
We parameterize $p(a_t|z_t, s_t)$ (\cref{fig:graphical_model}, right)
using just the current state $s_t$, whereas the encoder $p(z_t|\Omega, (s_1, \cdots, s_t))$
uses all previous states since 
a sequence of state observations $(s_1, \cdots, s_t)$ could potentially be very
informative of the $\Omega$ being followed, which if provided directly to $p(a_t|\cdot)$ 
can lead to a leakage of the option information 
to the actions, rendering the bottleneck on option information imposed via $z_t$ useless.}
Hence, in our implementation we remove recurrence over partial observations
for $p(a_t| z_t, s_t)$ while keeping it in $p(z_t\mid \Omega, s_t)$.}

\iftoggle{arxiv}{
\begin{algorithm}[t]
    \iftoggle{arxiv}{\scriptsize}{}
	\caption{\nirbhay{IR}-VIC}
	\begin{algorithmic}
		\label{algo:train_algo}
		\REQUIRE A parameterized encoder $p_{\phi}(z_t\mid\omega, \nirbhay{x_t})$, policy $\pi(a_t\mid \omega, \nirbhay{x_t})$
		\REQUIRE A parameterized option inference network $q_{\nu}(\omega \mid s_0, s_f)$
		\REQUIRE A parameterized goal-conditioned policy $\pi_{\gamma}(a_t|\nirbhay{x_t}, g)$
		\REQUIRE A prior on discrete options $p(\omega)=\rama{\frac{1}{|\Omega|}}$ \nmrebuttal{and integer $H$ - the length of each option trajectory.}
		\REQUIRE A variational approximation 
		of the option-marginalized encoder $q(z_t)$
		\REQUIRE A regularization weight $\beta$ and max-ent coefficient $\alpha$
		\REQUIRE A set of training environments $p_{\mathtt{train}}(E)$ and transfer environments $p_{\mathtt{transfer}}(E)$ \\
		\STATE \textbf{Unsupervised Discovery}
		\STATE Sample training environment $E_{\mathtt{train}} \sim p_{\mathtt{train}}(E)$
		\FOR{episodes = 1 to $\mathtt{max-episodes}$}
		\STATE Sample a spawn location $S_0 \sim p(s_0|E_{\mathtt{train}})$ and an option $\Omega \sim p(\omega)$
		\STATE Unroll a state-action trajectory $\tau$ under $\pi_{\theta}(a_t|\nirbhay{x_t}, z_t)$ for \nmrebuttal{$H$} steps with reparametrized $Z_t \sim p_{\phi}(z_t | \nirbhay{x_t}, \omega)$
		\STATE Infer $\Omega$ from $q_{\nu} (\omega | s_o, s_f)$
		\STATE Update the parameters $\theta,\nu$ and $\phi$ based on Eqn.~\ref{eq:usd_final_obj}
		\ENDFOR
		\STATE \textbf{Transfer to Goal-Driven Tasks}
		\STATE Sample transfer environment $E_{\mathtt{transfer}} \sim p_{\mathtt{transfer}}(E)$
		\FOR{episodes = 1 to $\mathtt{max-episodes}$}
		\STATE Sample a goal $G \sim p(g | E_{\mathtt{transfer}})$
		\STATE Unroll a state-action trajectory under the goal-conditioned policy $\pi_{\gamma}(a_t|\nirbhay{x_t}, g)$
		\STATE Update policy parameters $\gamma$ to maximize the reward given by Eqn.~\ref{eq:bonus_reward}
		\ENDFOR
	\end{algorithmic}
\end{algorithm}
}{}

%% file: new_sections/results.tex
\label{sec:results}
\paragraph{\textbf{Environments.}}
We pre-train and test on grid-worlds from
the MiniGrid~\citep{gym_minigrid} environments.
We first consider a set of simple environments
-- \texttt{4-Room} and \texttt{Maze}
(see Fig.~\ref{fig:prelim_env_dec_states}) followed by the \texttt{MultiRoomNXSY}
also used by~\citet{goyal2019infobot}.
The \texttt{MultiRoomNXSY} environments consist of
X rooms of size Y, connected in random orientations. \nirbhay{We refer to the ordering
of rooms, doors and goal as a `layout' in the \texttt{MultiRoomNXSY} environment
-- pre-training of options (for \irvic and \diayn) is performed
on a single fixed layout while
transfer is performed on several different layouts (a layout is randomly
selected from a set every time the environment is reset).}
In all pre-training environments,
\nmcr{we fix the option trajectory length $H$ (the number of steps
an option takes before termination) to 30 steps.}

We use Advantage Actor-Critic (A2C) for all experiments.
Since code for \ibot~\citep{goyal2019infobot}
was not public, we report numbers based on a
re-implementation of \ibot, ensuring consistency with
their architectural and hyperparameter choices.
\nmcr{We refer the readers to our
code\footnote{\url{https://github.com/nirbhayjm/irvic}}
for further details.}

\paragraph{\textbf{Baselines.}} We evaluate the following on quality of exploration and transfer to downstream goal-driven
tasks with sparse rewards:
    1) \ibot (our implementation) -- which identifies goal-driven
decision states by regularizing goal information,
    2) \nirbhay{\diayn~-- whose focus is unsupervised
skill acquisition,}
but has an $I(A_t, \Omega| S_t)$ term which can be used
for the bonus in~\Eqref{eq:bonus_reward},
    3) count-based exploration which uses visitation counts
as exploration incentive (this corresponds to replacing
$I(\Omega, Z_t | S_t, S_0)$ with $1$ in~\Eqref{eq:bonus_reward}),
    4) a randomly initialized encoder $p(z_t\mid\omega, x_t)$
    \nirbhay{which is a noisy version of the count-based baseline
    where the scale of the reward is adjusted to match the
    count-based baseline}
    5) how different values of $\beta$ affect performance and how we
    choose a $\beta$ value using a validation set, and
\nirbhay{
    6) a heuristic baseline that uses domain knowledge to
identify landmarks such as corners and doorways and provide a higher
count-based exploration bonus to these states.
} This validates the extent to which
\nirbhay{necessary option information is useful in identifying
a sparse set of states that are useful for transfer
\vs heuristically determined landmarks.}

\subsection{Qualitative Results}
\Figref{fig:prelim_env_dec_states} shows \nirbhay{
heatmaps of necessary option information
$I(\Omega, A_t | S_t, S_0)$} on \fourroom and \maze grid world environments
\nirbhay{where the initial state is sampled uniformly at random.}
\nirbhay{Stronger regularization ($\beta=1$) leads to poorer empowerment
maximization and in some cases not learning any options
(and $I(\Omega, A_t | S_t, S_0)$ collapses to 0 at all states).
}
At lower values of $\beta=\texttt{1e-3}$, we get more discernible
\nirbhay{states with distinctly high necessary option information.}
Finally, for \maze we see that for a similar
value of empowerment\footnote{Since \diayn maximizes the mutual
information with every state in a trajectory, we report the
empowerment for the state with maximum mutual information
with the option.}, \nirbhay{\irvic leads to a more peaky distribution of states
with high necessary option information} than \diayn.

\newcommand{\nocom}{}
\begin{table}[t]
	\resizebox{0.480\textwidth}{!}{
		\begin{tabular}{lccc}
			\toprule
			\textbf{ Method} &  \textbf{MR-N3S4} &   \textbf{MR-N5S4} & \textbf{MR-N6S25}\\
			\midrule
			$p_{\phi}(Z_t|S_t, \Omega)$ pretrained on & MR-N2S6 & MR-N2S6 & MR-N2S10 \\
			\midrule
			\ibot~\citep{goyal2019infobot}     			& $90\%$ & $85\%$ & {\tt N/A}  \\
			\midrule
			\ibot (Our Implementation)     				& $99.9\%$\nocom{$\pm0.1\%$} & $79.1\%$\nocom{$\pm11.6\%$} & $90.9\%$\nocom{$\pm1.2\%$}  		  \\
			Count-based Baseline                         & $99.7\%$\nocom{$\pm0.1\%$} & $99.7\%$\nocom{$\pm0.1\%$} & $86.8$\%\nocom{$\pm2.2\%$}   		  \\
			DIAYN                                           & $99.7\%$\nocom{$\pm0.1\%$} & ${95.4}\%$\nocom{$\pm4.1\%$} &  $0.1\%$\nocom{$\pm0.1\%$}   		  \\
			Random Network     				& ${99.9}\%$\nocom{$\pm0.1\%$} & $98.8\%$\nocom{$\pm0.7\%$} & $79.5\%$\nocom{$\pm5.2\%$}  \\
			Heuristic Baseline         					& {\tt N/A} & {\tt N/A} & $85.9\%$\nocom{$\pm3.0\%$}  \\
			Ours ($\beta=10^{-2}$)     					& ${99.3}\%$\nocom{$\pm0.3\%$} & $99.4\%$\nocom{$\pm0.2\%$} & ${92.9}\%$\nocom{$\pm1.2\%$}  		  \\
			\bottomrule
		\end{tabular}
	}
	\caption{
	Success rate (mean $\pm$ standard error) of the goal-conditioned policy when
	trained with different exploration bonuses in addition to the extrinsic
	reward $R_e(t)$. We report results at $5\times10^5$ timesteps for
	MultiRoom (MR-NXSY) N3S4, N5S4 and at $10^7$ timesteps for
	N6S25. We also report the performance of \ibot for completeness.
	Note that for rooms of size 4 (N3S4, N5S4),
	incentivizing to visit corners and doorways (Heuristic Sub-goals)
	is equivalent to count-based exploration.}
	\label{tab:transf_success_rate}
\end{table}

\subsection{Quantitative Results}
\subsubsection{Transfer to Goal-driven Tasks}
Next, we evaluate \Eqref{eq:bonus_reward}, \ie whether
\nirbhay{providing visitation incentive proportional
to necessary option information at a state}
in addition to
\nirbhay{sparse} extrinsic reward can aid in transfer
to goal-driven tasks in different environments.
We restrict ourselves to
the point-navigation task~\citep{goyal2019infobot}
transfer in the \texttt{MultiRoomNXSY} set of
partially-observable
environments. In this task, the agent
\nirbhay{learns a policy $\pi(a_t | g_t, s_t)$ where
$g_t$ is the vector pointing to the goal
from agent's current location at every time step $t$.}
\nirbhay{The initial state is always}
the first room, and has to go to a randomly sampled goal location
in the last room
and is rewarded only when it reaches the \nirbhay{goal}.~\citet{goyal2019infobot}
test the efficacy of different exploration objectives~
\footnote{
    While our focus is on identifying and probing how good
    sub-goals from intrinsic training are,
    more broader comparisons to exploration baselines are
    in \ibot~\citep{goyal2019infobot}.
}
and show that this is a hard
setting where efficient exploration is necessary.
They show that \ibot
outperforms several state-of-the art exploration
methods in this environment.

Concretely, we 1) train
\irvic to identify sub-goals (\Eqref{eqn:usd_objective}) on
\texttt{MultiRoomN2S6} and transfer to a
goal-driven task on \texttt{MultiRoomN3S4}
and \texttt{MultiRoomN5S4} (similar to~\cite{goyal2019infobot}),
and 2) train on \texttt{MultiRoomN2S10} and transfer to
\texttt{MultiRoomN6S25}, which is a more challenging transfer
task \ie it has a \nirbhay{larger upper limit on room size}
making efficient exploration critical to find doors quickly.
\nirbhay{
}
\nirbhay{For \irvic and \diayn (the two methods that learn options),
we pre-train on a single layout of the corresponding
\texttt{MultiRoom} environment for $10^6$ episodes
and pick the checkpoints with highest empowerment values across training.
For \ibot (no option learning required), we pre-train as per
\citet{goyal2019infobot} on multiple layouts of the \texttt{MultiRoom}
environment. Transfer performance of all methods
is reported on a fixed test set of
multiple \texttt{MultiRoom} environment layouts and \rama{hyperparameters
across all methods, \eg $\beta$ for \irvic and \ibot} are selected
using a validation set of \texttt{MultiRoom}
environment layouts.}
\begin{figure}[t!]
	\centering
	\includegraphics[width=0.37\textwidth]{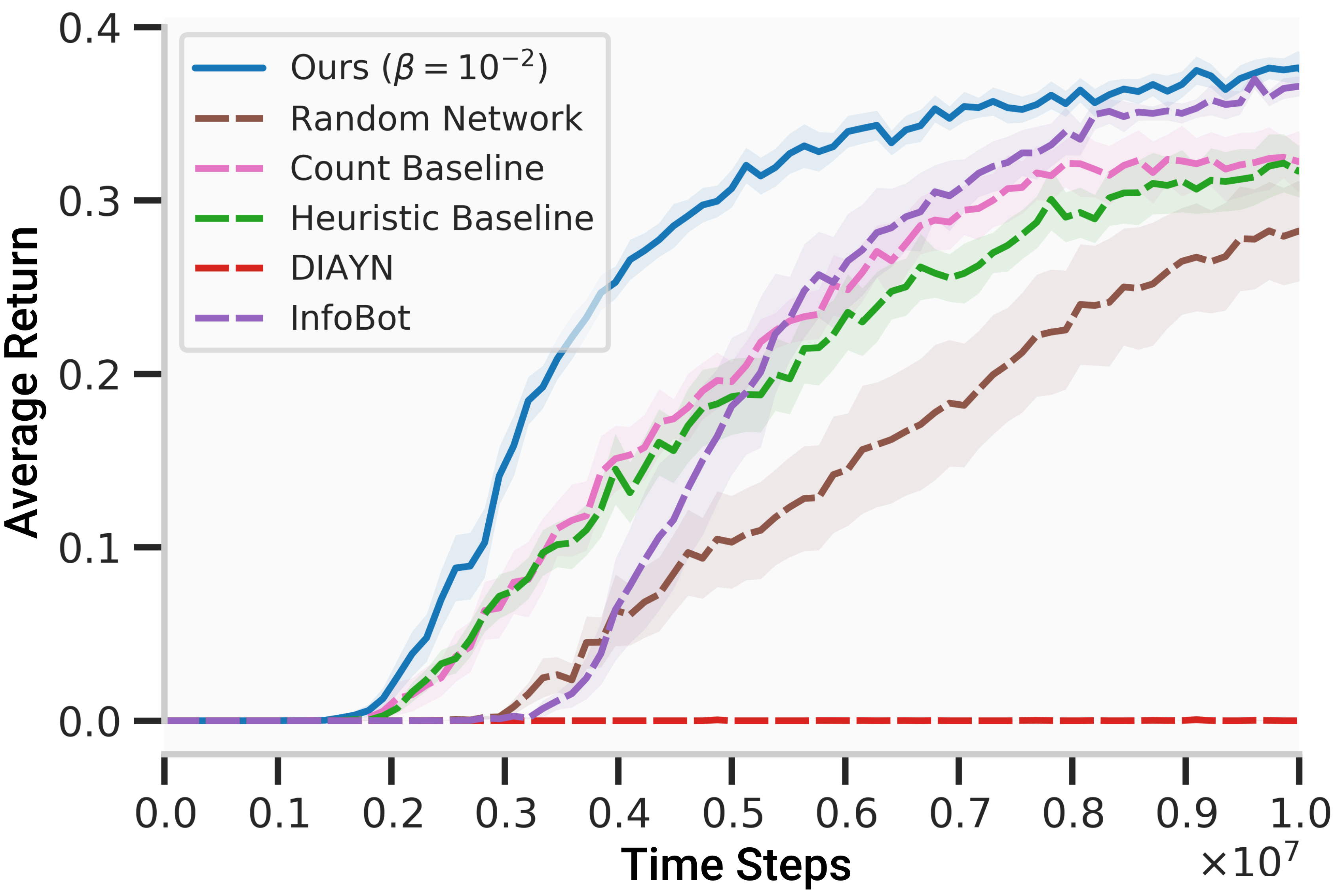}
	\caption{
	Transfer results on a test set of MultiRoomN6S25 environment layouts
	after unsupervised pre-training on \nirbhay{MultiRoomN2S10}.
	\nmcr{Shaded regions represent standard error of the mean of
	average return over 10 random seeds}.}%
	\label{fig:sample_complex_viz}%
\end{figure}
\subsubsection{Overall Trends}
Table~\ref{tab:transf_success_rate} reports
success rate -- the \% of times the agent reaches the goal and
\nirbhay{\Figref{fig:sample_complex_viz} reports the average return \rama{when
learning to navigate on test environments}}.
\nirbhay{The \texttt{MultiRoomN6S25} environment provides a sparse decaying
reward upon reaching the goal -- implying that when comparing methods,
higher success rate (Table~\ref{tab:transf_success_rate})
indicates that the goal is reached more often,
and higher return values (\Figref{fig:sample_complex_viz}) indicate that
the goal is reached with fewer time steps.}

First, our implementation of \ibot is competitive with~\citet{goyal2019infobot}\footnote{We found it important
to run all models (inlcuding \ibot) an order of
magnitude more steps compared to
~\citet{goyal2019infobot},
but our models also appear to converge
to higher success values.}.
Next, for the MultiRoomN2S6 to N5S4 transfer (middle column),
baselines as well as sub-goal identification methods perform well
with some models having \nmcr{overlapping confidence intervals
despite low success means}.
In MultiRoomN2S10 to N6S25 transfer, where the latter has a large
state space, we find that \irvic (at $\beta = 10^{-2}$) achieves
the best sample complexity (in terms of average return) and
final success, followed closely by \ibot.
\nirbhay{Moreover, we find that the heuristic baseline which identifies
a sparse set of landmarks (to mimic sub-goals) does not perform well --
indicating that it is not easy to hand-specify sub-goals that are useful
for the given transfer task.}
Finally, the randomly initialized encoder as well as DIAYN
generalize much worse in this transfer task.

\begin{figure}[t]
	\centering
	\includegraphics[width=0.37\textwidth]{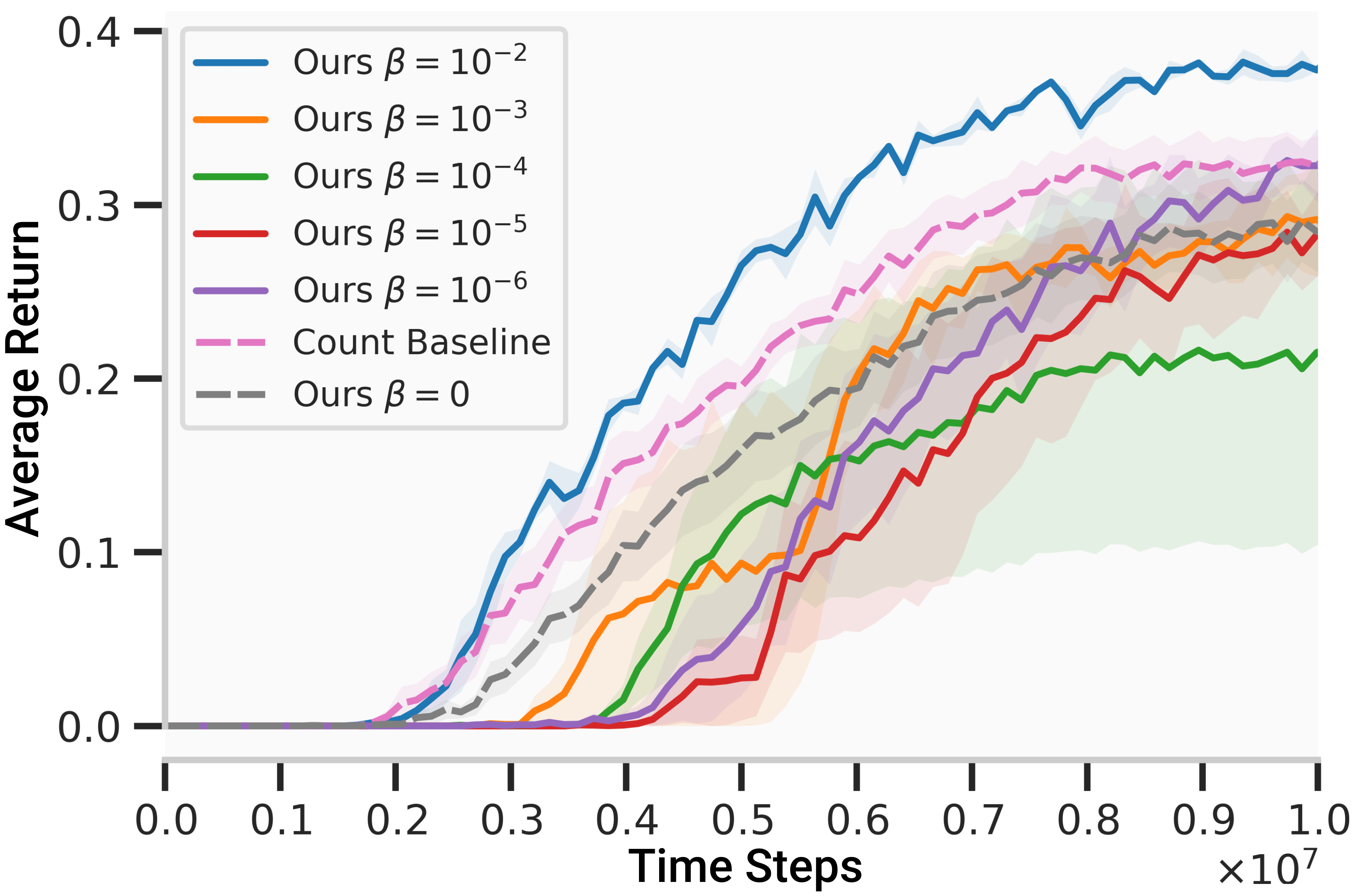}
	\caption{
	\nirbhay{Evaluation of average return on a held-out validation set of
	\texttt{MultiRoomN6S25} environment layouts.
	For each value of $\beta$, pre-traning is performed over 3 random seeds with
	the best seed being picked to measure transfer performance over 3 subsequent
	random seeds. Shaded regions represent standard error of the mean
	over the 3 random seeds used for transfer.
	}
    }
	\label{fig:beta_plot}
\end{figure}

\paragraph{\textbf{$\beta$ sensitivity.}}
We sweep over $\beta$
in log-scale from $\{10^{-1}, \dots, 10^{-6}\}$,
as shown in~\Figref{fig:beta_plot} (except $\beta=10^{-1}$ which does
not converge to $> 0$ empowerment) \nmrebuttal{and also report $\beta=0$ which recovers
a no information regularization baseline.}
\nirbhay{We find that $10^{-2}$ works best -- \rama{with performance tailing off
at lesser values. This is intuitive, since for a really large value of $\beta$,
one does not learn any options (as the empowerment is too low), while for a really
small value of $\beta$, one might not be able to target
necessary option information, getting large ``sufficient'' (but not necessary)
option information for the underlying option-conditioned policy.}}

\rama{We pick the best model for transfer based on performance on the validation
environments, and study generalization to novel test environments. Choosing the
value of $\beta$ in this setting is thus akin to model selection. Such design
choices are inherent in general in unsupervised representation
learning (\eg with K-means and $\beta$-VAE \cite{Higgins2017}).}

%% file: new_sections/related_work.tex
\paragraph{\textbf{Intrinsic control and intrinsic motivation.}} Learning
how to explore without extrinsic rewards is a foundational problem in
Reinforcement Learning~\citep{pathak2017curiosity,gregor2016variational,Schmidhuber1990}.
Typical curiosity-driven approaches attempt to visit states that maximize the
surprise of an agent~\citep{pathak2017curiosity} or improvement
in predictions from a dynamics model~\citep{lopes.t.m.p:2012:ExpMBEmpiricalError}.
While curiosity-driven approaches seek out
and explore novel states, they typically do not measure how
reliably the agent can reach them.
In contrast, approaches for intrinsic
control~\citep{eysenbach2018diversity,achiam2018variational,gregor2016variational}
explore novel states while ensuring those states are reliably
reachable.
\citet{gregor2016variational} maximize the number of final states
that can be reliably reached by the policy,
while \citet{eysenbach2018diversity}
distinguish an option (which they refer to as a `skill') at every state along the trajectory, and \citet{achiam2018variational}
learn options for entire trajectories by encoding a sub-sequence of states, sampled at regular intervals.
\rama{Since we wish to learn to identify useful sub-goals which one can reach}
reliably acting in an
environment rather than just visiting novel states (without
an estimate of reachability), we formulate our
regularizer in the intrinsic control framework, specifically building
on 
the work of~\citet{gregor2016variational}.

\paragraph{\textbf{Default behavior and decision states.}}
Recent work in policy compression has focused on learning a \textit{default policy}
when training on a family of tasks, to be able to re-use behavior across tasks.
In~\citet{teh2017distral}, default behavior is learnt
using a set of task-specific policies which then regularizes each policy, while~\citet{goyal2019infobot} learn a default policy using an information bottleneck on task information and a latent variable the policy conditions on,
\nirbhay{identifying sub-goals which they term as ``\decs''}.
We devise a similar information regularization objective
that learns default behavior shared by all intrinsic options
without external rewards so as to reduce learning pressure on option-conditioned policies.
\rama{Different from these previous approaches, our approach does not need
any explicit reward specification when learning options (ofcourse, since we
care about transfer we still need to do model selection based on validation
environments).}

\paragraph{\textbf{Bottleneck states in MDPs.}}
There is rich literature on identification of bottleneck states in
MDPs.
The core idea is to either identify
all the states that are common to multiple goals in an environment
~\citep{mcgovern_icml01} or use a diffusion model built using an MDP's
transition matrix~\citep{Machado2017}.
The key distinction between bottleneck states and \nirbhay{necessary-information} 
based sub-goals
is that the latter are more closely tied to the information
available to the agent and what it can act upon, whereas bottleneck
states are more tied to the connectivity structure of an 
MDP and intrinsic to the environment, representing
states which when visited allow access to a novel set of
states~\citep{goyal2019infobot}.
\nirbhay{However, bottleneck states do not easily apply to partially 
observed environments and when the transition dynamics of the MDP
are not known.}

\paragraph{\textbf{Information bottleneck in machine learning.}}
Since the foundational work of~\citet{Tishby99,Chechik2005},
there has been a lot of interest in making use of
ideas from information bottleneck (IB) for various
tasks such as clustering~\citep{Strouse2017c,Still2004},
sparse coding~\citep{Chalk2016-qp}, classification using
deep learning~\citep{Alemi2016},
cognitive science and language~\citep{Zaslavsky2018} and
reinforcement learning~\citep{goyal2019infobot,Strouse2018Share}.
We apply an information regularizer
to an RL agent that 
\nirbhay{results in a set of sparse states where necessary option 
information is high, which correspond to our sub-goals.}

%% file: new_sections/conclusion.tex
We devise a principled approach to identify \nirbhay{sub-goals}
in an environment without any extrinsic reward supervision using
a sandwich bound on the empowerment of~\citet{gregor2016variational}.
Our approach yields \nirbhay{sub-goals that}
aid \nirbhay{efficient} exploration on external-reward tasks and subsequently lead to
better success rate and sample complexity in novel environments
(competitive with supervised sub-goals).
Our code and environments will be made publicly available.

%% file: new_sections/appendix.tex
\subsection{Proof of \Cref{lemma:sandwich_bound}}\label{appendix:lemma_proof}
We state a proof of the \Cref{lemma:sandwich_bound} stating that
our proposed regularizer $\sum_{t=1}^{f} I(\Omega, A_t | S_t, S_0)$
forms an bound on empowerment $I(\Omega, S_f | S_0)$
from the main paper. This, combined with the lower-bound presented
in VIC~\citep{gregor2016variational}, forms a sandwich bound on 
$I(\Omega, S_f | S_0)$.

\textbf{Lemma 2.1} \prithvi{Let $A_t$ be the action random variable at timestep
$t$ and state $S_t$
following an option-conditioned policy $\pi(a_t | s_t, \omega)$.
Then, $I(\Omega, A_t | S_t, S_0)$ \ie the conditional mutual
information between the option $\Omega$ and action
$A_t$ when summed over all timesteps in the trajectory, upper bounds the
conditional mutual information $I(\Omega, S_f | S_0)$ between $\Omega$ and
the final state $S_f$ -- namely the empowerment as
defined by~\citet{gregor2016variational}:}
\begin{align*}
I(\Omega, S_f | S_0) \leq \sum_{t=1}^{f} I(\Omega, A_{t} | S_t, S_0) 
    = \mathcal{U}_{DS}(\vtau, \Omega, S_0)
\end{align*}

\par \noindent
\begin{proof}
\prithvi{To begin, observe that the graphical
model presented in Fig.~\ref{fig:graphical_model} satisfies the
markov chain $\Omega \leftrightarrow \{s_1, a_1\}_{t=1}^{f-1} \leftrightarrow S_f$ (assuming every node is conditioned on the intitial state $S_0$). Therefore, the data-processing inequality (DPI)~\cite{Cover91} implies:}
\begin{align}
    I(\Omega, S_f | S_0) \leq I(\Omega, (S_{f-1}, A_{f-1}) | S_0)\nonumber
\end{align}

\prithvi{Furthermore, using the chain rule of mutual information~\cite{Cover91}, we can write:}
\begin{align*}
I(\Omega, S_f | S_0) 
    &\leq I(\Omega, (S_{f-1}, A_{f-1}) | S_0) \\
    &= I(\Omega, S_{f-1} | S_0) 
    + I(\Omega, A_{f-1} | S_{f-1}, S_0)   
\end{align*}
\prithvi{Repeating the same set of steps recursively gives us:}
\begin{align*}
&I(\Omega, S_f | S_0)\\
    &\leq I(\Omega, (S_{f-1}, A_{f-1}) | S_0) \\
    &=I(\Omega, S_{f-1} | S_0) + I(\Omega, A_{f-1} | S_{f-1}, S_0)\\
    &=I(\Omega, S_{f-2} | S_0) + I(\Omega, A_{f-2} | S_{f-2}, S_0) \notag\\
    \uatext{}
    &\phantom{AAAAAAAAAA}+ I(\Omega, A_{f-1} | S_{f-1}, S_0) \\
    &\cdots\\
    &=I(\Omega, S_{0} | S_0) + \sum_{t=1}^{f} I(\Omega, A_{t} | S_t, S_0)
\end{align*}

Note that the graphical model presented in Fig.~\ref{fig:graphical_model} implies that $\Omega \independent S_0$ and hence,
\begin{align*}
    I(\Omega, S_{0} | S_0) = H(\Omega) - H(\Omega | S_0) = H(\Omega) - H(\Omega) = 0
\end{align*}
\begin{align*}
    \Rightarrow I(\Omega, S_f | S_0) \leq \sum_{t=1}^{f} I(\Omega, A_{t} | S_t, S_0) \phantom{AAAAAAAAAAAAAAA}
\end{align*}
\end{proof} 

\subsection{Upper bound on \texorpdfstring{$I(A_t, \Omega|S_t, S_0)$}{I(at,Omega|st,s0)}}
\label{sec:lower_bound_mi}
We explain the steps to derive Eqn. (4) in the main paper,
as an upper bound on $I(A_t, \Omega| S_t, S_0)$.
By the data processing inequality~\citep{Cover91}
$I(A_t, \Omega| S_t, S_0) \le I(Z_t, \Omega| S_t, S_0)$ for the
graphical model in Fig \ref{fig:graphical_model}. %
So we will next derive an upper bound
on $I(Z_t, \Omega| S_t, S_0)$.

We write $I(\Omega, Z_t| S_t, S_0=s)$, given a start state $S_0=s$ as:
\begin{align*}
\mathop{\E}\limits_{
    \substack{
        \Omega\sim p(\omega) \\
        S_t\sim p^{J}(s_t|\Omega, S_0=s) \\
        Z_t \sim p(z_t| S_t, \Omega)
    }
}
\left[\log \frac{p(Z_t|\Omega, S_t)}{p(Z_t|S_t)}\right]
\end{align*}
The key difference here is that our objective here uses the options $\Omega$
\emph{internal} to the agent as opposed to~\citet{goyal2019infobot}, who use
external goal specifications $g$ provided to the agent. Similar to VIC,
$p^J$ here denotes the (unkown) state distribution at time $t$ from which we can
draw samples when we execute a policy.

We then assume a variational approximation
$q(z_t)$\footnote{For our experiments,
we fix $q(z_t)$ to be a unit gaussian, however it could also be learned.}
for $p(z_t| S_t)$, and using the fact that
$\KL[p(z_t| s_t)|| q(z_t)] \ge 0$, we get a the
following lower bound:
\begin{align}
I(\Omega, Z_t| S_t, S_0=s) 
&\ge \mathop{\E}\limits_{
    \substack{
        \Omega \sim p(\omega)\\
        S_t \sim p^J(s_t| \Omega, S_0=s)\\
        Z_t \sim p(z_t| S_t, \Omega)
    }
}\left[
    \log \frac{p(Z_t|\Omega, S_t)}{q(Z_t)}
\right]
\label{appendix_eq:usd_local_bound}
\end{align}

\subsection{\irvic:
 On policy with Options.}\label{appendix:irvic_on_policy}
\rama{
We use Eqn.~\ref{appendix_eq:usd_local_bound} to 
\nmrebuttal{discover}
and visualize the
\decs learned in an environment, augmented with random sampling of the start state $S_0$.
Thus, we compute our decision states in an on-policy manner. Mathematically, we can write this
as:}

\begin{equation}
\mathop{\E}\limits_{ 
    \substack{
        \Omega \sim p(\omega),
        S_0 \sim p(s_0),\\ 
        S_t \sim p^J(s_t| \Omega, S_0),
        Z_t \sim p(z_t| S_t, \Omega)
    }
}\left[
    \log \frac{p(Z_t|\Omega, S_t)}{q(Z_t)}
\right]
\end{equation}
\rama{where $S_0$ is a random spawn location uniformly chosen from the set of states
in the environment and $\Omega$ is a random option chosen at each of the spawn locations.
Thus, for each state $S_t$ in the environment, we look at the aggregate of all the trajectories
that pass through it and compute the values of $\log \frac{p(z_t|\omega, s_t)}{q(z_t)}$ to
identify / visualize decision states. In addition to being principled, this
methodology also precisely captures our intuition that it is possible to identify decision
states which are common to, or frequented across multiple options. Our
results~\Cref{sec:results} show that the identified \decs match our assessments of \decs corresponding to some structural regularities 
in the environment.}

\subsection{Sub-goals for Transfer}
\label{appendix:subgoals_in_transfer}
As mentioned in the main paper, we would like to compute the MI: $I(\Omega, Z_t| S_t, S_0)$
to identify sub-goals as state where this MI value is high 
and given a (potentially) novel environment and a novel
goal-conditioned task, to provide this MI value as an exploration bonus. 
Given a state $s'$,
that we would like to compute $I(\Omega, Z_t| S_t=s', S_0)$, we can write:
\begin{align*}
\sum_{\omega, s_0}\int \ 
    &p(\omega) 
    p(s_0) 
    p^J(S_t=s'| s_0, \omega) 
    p(z_t| s_t, \omega) \\
    &\log \frac{p(z_t|\omega, S_t=s')}{q(z_t)}
    \ dz_t
\end{align*}
However, this cannot be computed in this form in a transfer task,
since a goal driven agent is not following on-policy actions for an option $\Omega$
that would allow us to draw samples from $p^J(\cdot| S_0, \Omega)$ above (in order to do a
monte-carlo estimate of the integral above). Thus instead, we propose to compute the
mutual information as follows:
\begin{align*}
\sum_{\omega, s_0}\int \ 
    &p(s_0, \omega|S_t=s') 
    p(z_t| S_t=s', \omega) \\
    &\log \frac{p(z_t|\omega, S_t=s')}{q(z_t)}
    \ dz_t
\end{align*}
Now, given a state $S_t=s'$, this requires us to draw samples from $p(s_0, \omega|S_t=s')$, which
in general is intractable (since this requires us to know $p^J(s_t| \Omega)$, which is
not available in closed form). In order to compute the above equation, we make
the assumption that $p(s_0, \omega| S_t=s') = p(s_0) p(\omega)$.
Breaking it down, this
makes two assumptions: firstly, $p(\omega | S_t=s', s_0) = p(\omega|s_0)$.
This means that all options have equal probability of passing through a state $s'$
at time $t$, which is not true in general. The second assumption is the same as VIC, namely
that $p(\omega | s_0) = p(\omega)$. Instead of making this assumption, one could
also fit a parameterized variational approximation to $p(\omega | S_t=s', s_0)$
and train it in conjunction with VIC. However, we found our simple scheme
to work well in practice, and thus avoid fitting an additional variational approximation.

\subsection{Baselines}
\nirbhay{
We use an exploration bonus coefficient of $\kappa = 0.1$ for the count-based exploration bonus baseline. The heuristic baseline identifies all occurrences of the following types of states in the MultiRoom environments -- (1) corners of the room, (2) doorways, and gives them a slightly higher coefficient of exploration bonus than the $\kappa$ used for the count-based bonus. We ran a sweep for the values of this higher coefficient and found that $0.105$ (i.e. a $+5\%$ increase) gave best results.
}

\subsection{Implementation Details}
\label{sec:implementation_det}
\par \noindent 
\textbf{Network Architecture:} 
We use a 3 layered convolutional neural network with kernels of size 3x3, 2x2 and 2x2 in the 3 layers respectively
to process the agent's egocentric observation.
We use \texttt{ReLU} as the non-linear activation function after each convolutional layer.
The output of the CNN is then concatenated with the agent's direction vector (compass) and
the option (or goal encoding).
The concatenated features are then passed through a linear layer with hidden size 64 to produce the final features used by option encoder
$p_{\phi}(z_t|s_t, \omega)$ head and the policy head $\pi_{\theta}(a_t|s_t, z_t)$. We use the (x, y) coordinates of the final state (embedded
through a linear layer of hidden size 64) to regress to the option via $q_{\nu}(\omega | s_0, s_f)$.
Furthermore, our parameterized policy is a reactive one and the encoder $p_{\phi}(z_t | s_t, \omega)$ is recurrent over the sequence of states encountered in the episode. The bottleneck random variable $Z_t$ is sampled from the parameterized gaussian $p_{\phi}(z_t | s_t, \omega)$ and is made a differentiable stochastic node using the re-parmaterization trick for gaussian random variables.

\par \noindent 
\textbf{Training Details:} We use Advantage Actor-Critic (A2C) (open-sourced implementation by~\citet{pytorchrl}%
) for all our experiments. We use RMSprop as the optimizer for all our experiments. \nirbhay{For the partially-observable grid world settings,} the agent receives an egocentric view of it's surroundings as input, encoded as an occupancy grid where the channel dimension specifies whether the agent or an obstacle is present at an $(x,y)$ location. We set the coefficient $\alpha$ in Eqn. \ref{eq:usd_final_obj} to be $10^{-3}$ for all our experiments based on sweeps conducted across multiple values. In practice, we found it to be difficult to optimize our unsupervised objective in absence of an entropy bonus -- the parameterized policy collapses to a deterministic one and no options are learned in addition to inefficient exploration of the state-space. Moreover, we found that it was hard to obtain reasonable results by optimizing for both the terms in the objective from scratch and therefore, we optimize the intrinsic objective itself for $\sim8k$ episodes (i.e., we set $\beta=0$) after which we turn on the regularization term and let $\beta$ grow linearly for another $\sim8k$ episodes to get feasible outcomes at convergence. We experiment with a vocabulary of 2, 4 and 32 options (imitating undercomplete and overcomplete settings) for all the environments. For all the exploration incentives presented in Table 1, we first picked a value of $\kappa$ in Eqn. \ref{eq:bonus_reward_infobot} (decides how to weigh the bonus with respect to the reward) from $\{10, 1, 0.1, 0.01, 0.001, 0.0001\}$ based on the best sample-complexity on goal-driven transfer.

\par \noindent
\textbf{InfoBot Implementation:} 
Since code to reproduce InfoBot~\citep{goyal2019infobot}
was not publiclt available,
we implemented InfoBot ourselves while making sure
that we are consistent with the architectural and hyper-parameter choices adopted by InfoBot (as per the A2C implementation by~\citep{gym_minigrid}).
(i) We replace the convolution layers with fully connected layers to process the observations, and (ii) we use the layer sizes as mentioned by \citep{goyal2019infobot} in the appendix. 
The checkpoint used to report the final performance for N3S4 and N5S4 were picked by doing 
\nirbhay{validation on the transfer task based on the success metric.}
However, this procedure was infeasible for the bigger N6S25 environment and we chose the last checkpoint from the training for this environment.

\par \noindent
\textbf{Convergence Criterion.}
In practice, we observe that it is hard to learn a large number of discrete options using the unsupervised objective. From the entire option vocabulary, the agent only learns a few number of options discriminatively (as identified by the option termination state), with the rest collapsing into the same set of final states. To pick a suitable checkpoint for the transfer experiments, we pick the ones which have learned the maximum number of options reliably -- as measured by the likelihood of the correct option from the final state $\log q_{\nu}(\omega | s_0, s_f)$.

\par \noindent
\textbf{Option Curriculum}
It is standard to have a fixed-sized discrete option space $\Omega$ with a uniform pior~\citep{gregor2016variational}.
However, learning a meaningful option space with larger option vocabulary size $|\Omega| = K$ has been reported to be difficult \cite{achiam2018variational}.
We adopt a curriculum based approach proposed in \cite{achiam2018variational} where the option vocabulary size is gradually increased as the option decoder $q_{\nu}(\omega | s_f)$
becomes more confident in mapping back the final state back to the corresponding option sampled at the beginning of the episode.
More concretely, whenever $q_{\nu}(\omega | s_0, s_f) > 0.75$ (threshold chosen via hyperparameter tuning), the option vocabulary size increases according to
	\[ K \leftarrow \min \bigg(  \; \text{int}\big(\;1.5 \times K + 1 \big), K_{max} \bigg) \]
For our experiments, we start from $K=2$ and terminate the curriculum when $ K = K_{max}=32$.

We run each of our experiments on a single GTX Titan-X GPU, and use no more than 1 GB of GPU memory per experimental setting.